\documentclass[letterpaper]{article}
\usepackage{aaai}
\usepackage{times}
\usepackage{helvet}
\usepackage{courier}
\usepackage{wrapfig}

\usepackage{amsthm}
\usepackage{balance}

\newtheorem{Definition}{Definition}

\usepackage{listings}
\usepackage{array}
\usepackage[ruled]{algorithm}
\usepackage[noend]{algpseudocode} 
\usepackage{todonotes}
\usepackage{rotating}
\usepackage{color}
\usepackage{color}
\usepackage{amsmath}
\usepackage{amssymb}

\newcommand{\code}[1]{{\texttt{\footnotesize #1}}}
\newcommand{\co}[1]{\code{#1}}

\renewcommand{\i}[1]{\textit{#1}}
\newcommand{\f}[1]{\textbf{#1}}

\newcommand{\itz}[1]{\begin{itemize} \addtolength{\parskip}{-1ex} #1 \end{itemize}}

\newcommand{\lam}[0]{\leftarrow}

\newcommand{\ol}[1]{\overline{#1}}

\newcommand{\eg}[0]{e.g.~}
\newcommand{\ie}[0]{i.e.~}

\newcommand{\wrt}[0]{wrt.~}

\newcommand{\citeaj}[1]{\citeauthor{#1} (\citeyear{#1})}

\lstdefinestyle{pddl}{
		basicstyle=\ttfamily\scriptsize,
		numbers=left, 
		stepnumber=1, 
		numberstyle=\tiny, 
		numbersep=10pt,
		xleftmargin=2em,
		frame=single,
}
\lstdefinestyle{lp}{
		basicstyle=\ttfamily\scriptsize,
		numbers=left, 
		stepnumber=1, 
		numberstyle=\tiny, 
		numbersep=10pt,
		xleftmargin=2em,
		frame=single,
}
\newcommand{\mc}[1]{\ensuremath{\mathcal{#1}}}

\definecolor{gray}{gray}{.5}  

\algrenewcommand\algorithmicthen{}
\algrenewcommand\algorithmicdo{}

\newcommand{\pws}[0]{\mc{PWS}}
\newcommand{\hpx}[0]{\mc{HPX}}
\newcommand{\Ak}[0]{$\mc{A}_k$ }

\newcommand{\menote}[1]{\footnote{\textcolor{blue!50!black}{\upshape Manfred: #1}}}

\renewcommand{\rm}[1]{\textcolor{orange}{#1}} 

\begin{document}
 \color{black}
%


\title{Narrative based Postdictive Reasoning for Cognitive Robotics
}



\author{Manfred Eppe and Mehul Bhatt \\
University of Bremen\\
{\small\{meppe, bhatt\}@informatik.uni-bremen.de}
}

\maketitle
 
\begin{abstract}

Making sense of incomplete and conflicting narrative knowledge in the presence of abnormalities, unobservable processes, and other real world considerations is a challenge and crucial requirement for  cognitive robotics systems. An added challenge, even when suitably specialised action languages and reasoning systems exist, is practical integration and application within large-scale robot control frameworks.

In the backdrop of an autonomous wheelchair robot control task, we report on application-driven work to realise postdiction triggered abnormality detection and re-planning for real-time robot control: (a) Narrative-based knowledge about the environment is obtained via a larger smart environment framework; and (b) abnormalities are postdicted from stable-models of an answer-set program corresponding to the robot's epistemic model. The overall reasoning is performed in the context of an approximate epistemic action theory based planner implemented via a translation to answer-set programming. 
\end{abstract}
%
%
%

\section{Introduction}
Researchers in the field of reasoning about action and change have interpreted \emph{narratives} in several ways, differing in the richness of their semantic characterisation and ensuing formal properties \cite{MillerS94,OccurNarraSC:Pinto:1998},\cite{DBLP:journals/lalc/Mueller07},\cite{McCarthy-98-kr-combining-narrative,McCarthy:2000:concepts-logical-AI}. For instance, within the context of formalisms such as the situation calculus and event calculus, narratives are interpreted as ``\emph{a sequence of events about which we may have incomplete, conflicting or incorrect information}'' \cite{MillerS94,OccurNarraSC:Pinto:1998}. The interpretation of narrative knowledge in this paper is based on these characterisations, especially in regard to the representation and reasoning tasks that accrue whilst modelling the perceptually grounded, narrativised epistemic state of an autonomous agent. In this paper, we are especially concerned with large-scale cognitive robotics systems where high-level symbolic planning and control constitutes one of many AI sub-components guiding low-level control and attention tasks \cite{Suchan-Bhatt-expcog-2012}.

\subsection{Perceptual Narratives and Postdiction}
Perceptual narratives correspond to declarative models of visuo-spatial, auditory, haptic and other observations in the real world that are obtained via artificial sensors and / or human input \cite{CMN-bhatt-et-al-2013}. From the formal viewpoint of commonsense reasoning, computational modelling and reasoning with perceptual narratives encompasses logics of  \emph{space, actions, and change} \cite{Bhatt:RSAC:2012}.\footnote{This paper does not directly address spatial representation and reasoning. Instead, the focus here is on \emph{action and change}.}

Declarative models of perceptual narratives can be used for interpretation, plan generation, and control tasks in the course of assistive technologies  in everyday life and work scenarios, e.g., in domains such as human activity recognition, semantic model generation from video, ambient intelligence and smart environments (e.g., see narrative based models in \cite{DBLP:journals/corr/abs-1202-3728,DBLP:conf/aaaiss/HajishirziM11,DBLP:journals/lalc/Mueller07,Bhatt:STeDy:10,dubba-bhatt-2011,CMN-bhatt-et-al-2013}). The focus of this paper is on particular inference patterns and an overall control architecture for online / incremental reasoning with narrative knowledge from the viewpoint of plan generation and explanation. We are especially interested in \emph{completion of narrative knowledge} by inferring perceptual  abnormalities and causes of perceived changes in the agent's world.


Explanation by postdictive reasoning within the framework of perceptual narratives can be the basis of explaining phenomena or properties perceived via sensory devices \cite{Poole:87,MillerS94}. Given perceptual narratives available as sensory observations from the real execution of a system, the objective is often to  assimilate / explain them with respect to an underlying domain / process model and an approach to derive explanations. The \emph{abductive explanation} problem can be stated as follows~\cite{kakas1992abductive}: given theory $T$, observations $G$, find an explanation $\bigtriangleup$ such that: (a). $T \bigcup \bigtriangleup \vDash G$; and  (b). $T \bigcup \bigtriangleup$ is consistent. In other words, the observation follows logically from the theory extended given the explanation. Amongst other things, this can be used to identify abnormalities in a narrative, which may in turn affect subsequent planning and overall (agent) control behaviour.



\subsection{Narrative-based Incremental Robot Control}
Our application of narrative-based incremental agent control is based on plan \i{monitoring}, and combining it with a mechanism for \emph{explanatory reasoning}: if a monitored world property changed unexpectedly, then our system postdicts possible explanations that describe what may have happened that caused this change. A common paradigm used in the planning community is \i{strong} planning, see \eg \cite{Bertoli2002}.  A strong plan guarantees that the goal is achieved, no matter how the (partially unknown) world is.  
However, this paradigm is not appropriate when considering abnormalities: it may always happen that a plan does not succeed due to unexpected system failures. As such, we use an \i{incremental weak planning} approach, and interleave the planning with plan execution. A weak plan must not guarantee that the goal will be achieved, it must only show possibilities to achieve a goal. The overall system is implemented such that as soon as one weak plan is found the system starts acting. This weak plan is then extended (\ie made ``stonger'') during plan-execution. Further, sensing results which are obtained during plan-execution are integrated in an online manner, and the search space is pruned accordingly during plan execution.

The narrative-driven explanation and control framework of this paper is built on a planning formalism called \i{h-approximation} (\hpx) that is incomplete but sound \wrt the possible-world semantics of knowledge \cite{Eppe2013b}. For \hpx\, a corresponding planning system has been implemented via translation to an answer-set program. The paper extends this planning system with new features such that:
\itz{
	\item it is capable of incremental online-planning, and
	\item it allows for abductive explanatory reasoning during plan execution. 
}


The paper presents an overview of the basic offline h-approximation, and describes the extensions of the proposed online version together with a detailed architecture of the overall control approach.\footnote{The extended planning system uses the online ASP reasoner \i{oclingo} \cite{Gebser2011b} that dynamically adopts its knowledge base according to sensing results. } We also bring forth the application guided motivations of our work by illustrating a real-time control task involving an autonomous wheelchair robot in a smart home environment. Finally, we present ongoing work aimed at integrating and delivering our online planner as a part of the experimental cognitive robotics framework ExpCog \cite{Suchan-Bhatt-expcog-2012}.

\section{Approximate Epistemic Planning as ASP}\label{sec:prelim}
We choose \hpx\ as the theoretical foundation for our framework because it has native and elaboration tolerant support for postdictive reasoning along with a low computational complexity (the plan existence problem is in NP). 
For alternative \pws\ based formalisms, plan existence is $\Sigma_2^P$-complete, \eg \cite{Baral2000}. 
To the best of our knowledge, no other implemented formalism supports postdiction in this complexity class.
The support for postdiction is crucial to realize abnormality- and explanation-based error-tolerance in robotic systems: 
if sensing reveals that the effect of an action is not as intended, then postdiction can be used to abduce and explain the reason for the failure. 
This is a partial solution to the \i{qualification problem}: it is not possible to model all conditions under which an action has the intended effect.
In this work we perform abnormality reasoning in an epistemic open-world sense. 
That is, we do not further describe (and circumscribe) abnormalities but rather use generic abnormality predicates as qualifications (negative conditions) of actions. 

\hpx\ is formalized and implemented in ASP: a problem specification is specified in PDDL-like syntax and is then translated to an Answer-Set Program \cite{Gelfond1988} via a set of translation rules.
The stable models of the generated Logic Program can be interpreted as conditional plans. 
The fact that \hpx\ is implemented as ASP and not in a procedural programming language like c++ makes it simple to extend the formalism and its semantics on a logical level. For this paper, ASP solvers like \i{oclingo} \cite{Gebser2011b} providing incremental and online problem solving-capabilities are relevant. Online problem solving makes it possible to dynamically add rules to a Logic Program. That is, the solver is in running in a loop and constantly awaiting new input via extra logical means. 
Whenever new rules are received, the solver tries to find new stable models according to the updated program. 

%


\f{Planning Problem Specification. \quad}
A problem domain is specified using a syntax similar to the planning domain definition language (PDDL): \co{(:init $l^{init}$)} represents initial knowledge about a literal $l^{init}$. 
\co{(oneof $l^{oo}_1 \hdots l^{oo}_n$)} describes initial exclusive-or-knowledge.
\co{(:action $a$ :effect when (and $l^c_1 \hdots l^c_n$) $l^e$)} is called an \i{effect proposition} (EP). 
It represents the conditional effect of an action $a$, that if the condition literals $l^c_1 \hdots l^c_n$ hold, then the effect $l^e$ will also hold.
\co{(:action $a$ :observe $f$)} is a \i{knowledge proposition}. 
It represents that an action $a$ will sense the value of a fluent $f$.
\co{(:action $a$ :executable (and$\,\,l^{ex}_1\hdots l^{ex}_n$))} is an \i{executability condition}. 
An action is only executable if literals $\,\,l^{ex}_1\hdots l^{ex}_n$ are known to hold.
Finally, \co{(:goal weak $l^{g}$)} is used to state weak goals, \ie goals which are satisfied by a plan such that a desired property $l^{g}$ is known to be achieved in at least one leaf of the search tree. 
We do not consider strong goals (a goal that must be known to hold in all leafs) because we consider an open world where it is impossible to model all qualifications of an action. Hence an action can always fail, and it is impossible to predict that a goal is achieved in all leafs of a transition tree.


\f{Planning Problem Formalization. \quad}
The h-approximation for a planning problem \mc{P} consists of two parts:
\itz{
\item $\Sigma_{hapx}$: a set or rules representing a foundational domain-independent theory
\item \mc{P}  {\small$\stackrel{\f{T}}{\longmapsto} \Sigma_{world}$}: the translation of a planning problem \mc{P} into a domain-specific theory $\Sigma_{world}$ using a set of translation rules \f{T}.
}
The resulting Logic Program, denoted by LP(\mc{P}), is the conjunction $\Sigma_{hapx} \cup \Sigma_{world}$.

The following are the main predicates used in the ASP formalization of the \hpx:
\itz{
\item $\mathit{occ}(a,t,b)$ denotes that action $a$ occurs at step $t$ in branch $b$. 
\item $\mathit{apply}(ep,t,b)$ denotes that an effect proposition $ep$ is applied at step $t$ in branch $b$.%
\footnote{In \hpx, actions are partitioned in EP to simplify reasoning with concurrency. Whenever $\mathit{occ}(a,t,b)$ and $ep$ is an effect proposition of $a$, then $\mathit{apply}(ep,t,b)$} 
\item $\mathit{sRes}(l,t,b)$ denotes that the literal $l$ is potentially sensed at step $t$ in branch $b$.
\item $\mathit{knows}(l,t,t',b)$ states that at step $t'$ in branch $b$ it is known that $l$ holds (or did hold) at step $t$ (with $t \leq t'$). That is, \hpx does not only consider knowledge about the present state but also about the past. 
\item $\mathit{nextBr}(t,b,b')$ denotes that sensing happened at step $t$ in branch $b$, resulting in a child-branch $b'$. 
\item $\mathit{uBr}(t,b)$ denotes that branch $b$ is a valid branch at step $t$. Actions can only be executed when a branch is valid.
\item $\mathit{goal}(l)$ denotes a (weak) goal for a literal $l$. 
}

\begin{figure}[!t]
\vspace{-22pt}
\hspace{-10pt}\includegraphics[width=1.05\columnwidth]{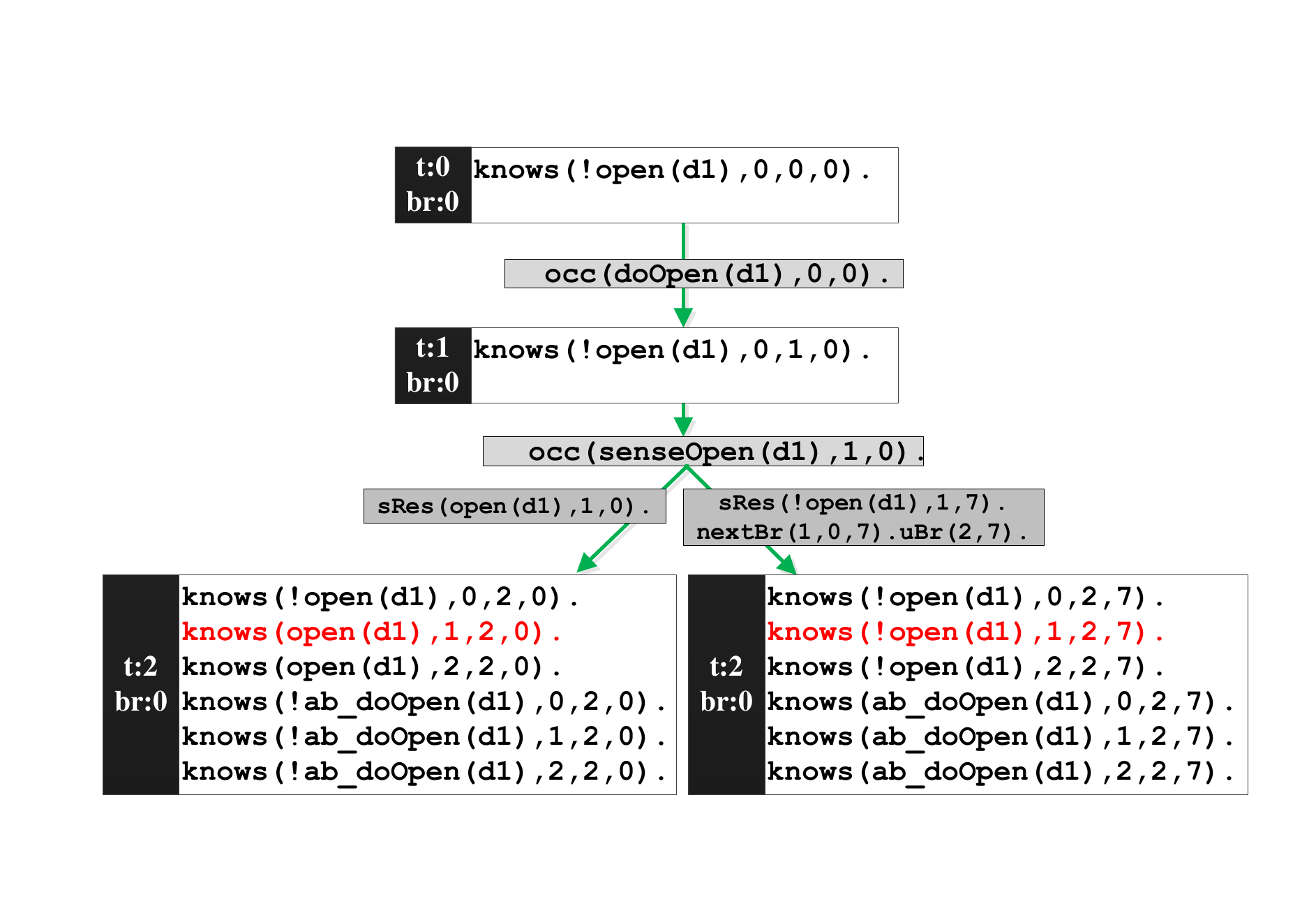}
\vspace{-42pt}
\caption{Example transition tree for \hpx}
\vspace{-5pt}
\label{fig:transTreeSimple}
\end{figure}
\begin{sloppypar}
\noindent \f{Example:\quad }
{
Consider the following action specification:
\begin{lstlisting}
(:action doOpen :parameters (?d - Door) 
         :effect when $\neg$ab_doOpen (open ?d))
\end{lstlisting}
}
This represents an action where a door will be open if there is no abnormality. 
This specification is translated into an ASP formalization via the translation rules (T6a-c) presented in \cite{Eppe2013b}. 
For instance, translation rule (T6a) generates:
\begin{equation*}
\begin{aligned}
\mathit{knows}&(open(D),T+1,T1,BR) \lam \\
&\mathit{apply(doOpen(D)_0,T,BR),} \\
&\mathit{knows(\neg ab\_doOpen(D),T,T1,BR),T1>T.}
\end{aligned}
\end{equation*}
That is, if the 0-th EP of the action \co{doOpen(D)} is applied at \co{T}, and at \co{T1}$>$\co{T} it is known that at \co{T} there is no abnormality then at \co{T1} it is known that after the action occurrence (at \co{T+1}) the door is open.
Similarly, translation rule (T6c) in \cite{Eppe2013b} generates:
\begin{equation*}
\begin{aligned}
&\mathit{knows(ab\_doOpen(D),T,T1,BR)} \lam \\
&~~~\mathit{knows( \neg open(D),T+1,T1,BR),}\\
&~~~\mathit{apply(doOpen(D)_0,T,BR),T1>T.}
\end{aligned}
\end{equation*}
 This reflects \i{negative postdiction}: If the open-action was executed at step \co{T} and at a step \co{T1} the effect of the open-action is known not to hold at \co{T+1}, then it is known that there must have been an abnormality at \co{T}. 
A similar \i{positive  postdiction} rule is generated by (T6b) which produces knowledge that there is no abnormality if it is known that the effect of the action holds after execution but not before execution. 

Consider a sensing action specification 
\begin{lstlisting}
(:action senseOpen :parameters (?d - Door) 
				 :observe (open ?d))
\end{lstlisting}
This generates knowledge as follows (see Figure \ref{fig:transTreeSimple}):
Assume the \co{drive}-action occurs in the initial node ($t=0$, $br=0$) in the transition tree. 
\co{senseOpen} is applied after \co{doOpen}, \ie at step 1 in branch 0: \ie $\mathit{occ(senseOpen(d1),1,0)}$. 
Then the positive sensing result is associated with the original branch 0: ($\mathit{sRes(open(d1),1,0)}$). The negative result is associated with a child branch, \eg 7:%
\footnote{The number of the child branch is randomly generated via a choice rule.}
 $\mathit{sRes(\neg open(d1),1,7)}$. 
Further, $\mathit{nextBr(1,0,7)}$ is generated to reflect that branch 7 is generated as a child of branch 0 at step 1. Finally, $\mathit{uBr(2,7)}$ is indicates that branch 7 is valid from step 2 on, and hence the planner may consider actions to occurr in branch 7. 

\end{sloppypar}

\section{Narrativised Online Robot Control}
The original h-approximation formalism and planning system is designed for offline problem solving. 
That is, a conditional plan is generated and the projected future world states are checked for whether a predefined goal holds. 
In this work, we extend h-approximation such that it is capable of online planning and abductive explanation. We also define several measures to assess the quality of a plan (e.g. robustness \wrt to unknown contingencies). A key feature of the h-approximation is the support for postdiction; we use postdiction to find explanations of why an action did or did not succeed. We propose to model actions such that the non-existence of an abnormality is a condition for the action to succeed.  After executing the action, sensing can be applied to verify whether an action succeeded, and thus to postdict whether there was an abnormality.  Identified abnormalities can then be used, for instance, in support of other kinds of reasoning or control tasks.

\subsection{I. An Extended Online h-approximation}
\begin{figure}
	\centering
		\includegraphics[width=0.9\columnwidth]{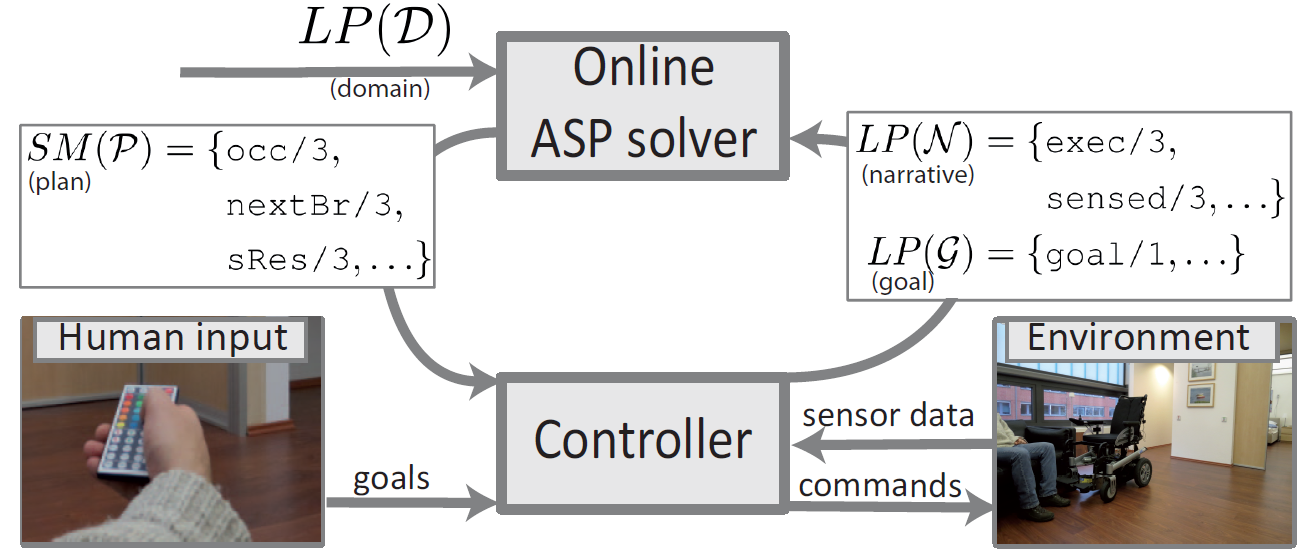}
	\caption{System Architecture}
	\label{fig:onlineASPtoolchain}
\end{figure}

The overall online h-approximation architecture is depicted in Fig. \ref{fig:onlineASPtoolchain}: 
it consists of an online ASP solver and a controller, which serves as interface to human input devices and the robotic environment. 

The complete LP to be solved is the conjunction of the LP translation of the domain specification \mc{D}, the goal specification \mc{G} and an execution narrative \mc{N}: $LP(\mc{P}) = LP(\mc{D}) \cup LP(\mc{G}) \cup LP(\mc{N})$.
Here, the execution narrative contains information about which actions were actually executed and which sensing results were actually obtained.


Once a stable model $SM(\mc{P})$ is found, it is sent to the controller which interprets this as a conditional plan and starts to execute it. 
It reports the execution narrative \mc{N} back to the solver.
The solver adopts the search space according to this information and refines / expands the plan accordingly. 
The updated stable models are thereupon sent to the controller again which acts accordingly.
The loop is repeated until the goal is achieved or the problem becomes unsolvable.

\begin{figure*}
		\hspace{-0.6cm}\includegraphics[width=1.07\textwidth]{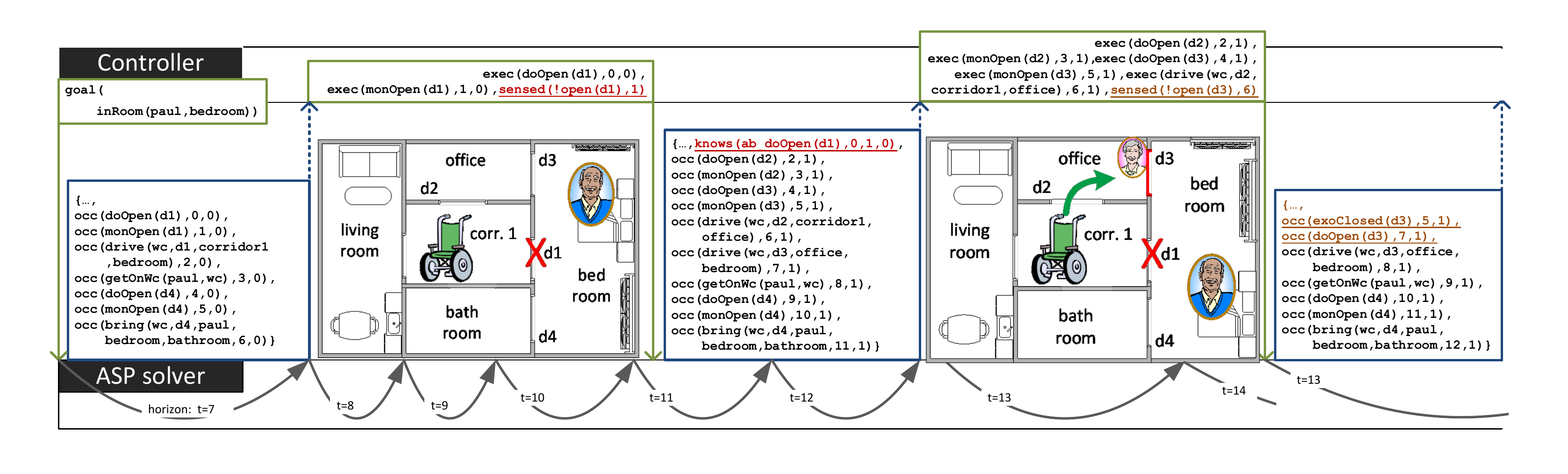}
	\caption{Communication between controller and ASP solver in Smart Home scenario\menote{TODO: remove the door between corr. 1 and bathroom.}}
	\label{fig:onlineCommunication}
\end{figure*}

\emph{Online Controller}.\quad We implement a controller which communicates new goals, sensing results and execution statements to the solver.
It is also responsible for the plan execution and the communication with actuators and sensors. 
Once an action is executed, the planner has to commit to this action, \ie it must always consider the occurrence of this action. 
This mechanism is implemented with the following rule:
\begin{equation}
\label{eq:commitActions}
\begin{aligned}
\mathit{occ}&(A,t,B) \lam \\
&\mathit{exec(A,t,B), a(A), uBr(t,B).}
\end{aligned}
\end{equation}
where $\mathit{exec}(a,t,b)$ represents that an action $a$ is executed at step $t$ in branch $b$. 
The controller sends this information to the solver when it starts to execute the action. 
Once an action is initiated its execution will not be aborted.

The following choice rule generates plans. 
{\small
\begin{equation}
\label{eq:genPlans}
\begin{aligned}
\{\mathit{occ(A,t,B)}& :a(A)\} \lam \\
&\mathit{uBr(t,B), notGoal(t,B),}\\
& \mathit{not~exec(A',t,B): a(A').}
\end{aligned}
\end{equation}
}
where $\mathit{notGoal}(t,b)$ denotes that the goal is not yet achieved in branch $b$ at step $t$. 
This prunes the search space because actions are only considered if the goal is not yet achieved at that node. 
Finally, $\mathit{not~exec(A',t,B)}$ causes the solver not to generate any actions at step $t$ if another action was already physically executed at that step.

Real-World sensing results are communicated from the controller to the solver in terms of $\mathit{sensed}$ atoms. 
These are integrated into the agent's knowledge state by disabling the effect of projected sensing results which do not coincide with the actual sensing:
{\small
\begin{align}
\label{eq:sResAssign}
&\mathit{sRes(F,t,B) \lam occ(A,t,B), hasKP(A,F),} \notag \\
&~~~~\mathit{not~knows(\neg F,t,t,B), not~sensed(\neg F,t).}\\
&~~~~\mathit{sRes(\neg F,t,B') \lam occ(A,t,B), hasKP(A,F),} \notag \\
&~~~~\mathit{not~kw(F,t,t,B), nextBr(t,B,B'), not~sensed(F,t).}
\end{align}
}
$\mathit{hasKP}(a,f)$ denotes that an action $a$ has a knowledge proposition  concerning a fluent $f$ (\ie it will sense $f$). 
Whenever such an action occurs, the positive sensing result is always projected to the original branch, while the negative result is projected on a child branch given through $\mathit{nextBr}$. 
Projected sensing results are only valid if they do not contradict the actual sensing results (implemented by $\mathit{not~sensed}$ statements).

Also, when receiving the actual sensing value, we have to take care that nodes which were valid in the projected search tree become invalid when the sensing contradicts the projections. 
The following rule implements that the original branch (where the pos. fluent was projected) becomes invalid if the sensing was negative ($\neg$ \co{F}) and the child branch becomes invalid if the sensing was positive:
{\small
\begin{equation}
\label{eq:sensingInvalid}
\begin{aligned}
brInvalid(t,B) &\lam sensed(\neg F,t),\\
occ(A,t,B)&, hasKP(A,F).\\
brInvalid(t,B') &\lam sensed(F,t), \\
occ(A,t,B)&, hasKP(A,F), nextBr(t,B,B').
\end{aligned}
\end{equation}
}

The information about invalid nodes is used for the generation of child branches. That is, a branch does not persist if it is invalid (\ref{eq:sensingRes1}) and it is also not created if invalid (\ref{eq:sensingRes2}).
{\small
\begin{subequations}
\label{eq:sensingRes}
\begin{equation}
\label{eq:sensingRes1}
uBr(t+1,B) \lam uBr(t,B), not~brInvalid(t,B).
\end{equation}
\begin{equation}
\label{eq:sensingRes2}
uBr(t+1,B') \lam nextBr(t,B,B'), not~brInvalid(t,B').
\end{equation}
\end{subequations}
}

\subsubsection{Incremental Planning Horizon Extension}
In online ASP solving, a single integer iterator (we use $t$) is incremented continuously until a solution is found.%
\footnote{Incremental problem solving is realized by splitting a LP up into three parts: $\mathit{\#base}$, $\mathit{\#cumulative}$ and $\mathit{\#volatile}$.  The $\mathit{\#base}$ part is an ordinary Logic Program while $\mathit{\#cumulative}$ and $\mathit{\#volatile}$contain the iterator which expands the problem horizon. With each incrementation a new ``slice'' of the Logic Program is grounded and added to the set of rules. Incrementation takes place until a solution is found or up to a certain limit, depending on the configuration of the solver.}
This is sensible for quickly finding a first solution for a planning problem, as it guarantees that if a plan is found then it is minimal in length. 
Also, this plan is usually found very quickly because the search space is relatively small in the beginning.

However, as we perform weak planning, it may well be the case that the first found plan does not lead to the goal in practice. Therefore the planning-horizon is constantly incremented while the robot is executing the plan. 
That is, the plan is expanded to consider more contingencies while the robot is already acting.

\noindent \f{Exogenous Actions} \quad (EA) are actions that occur but which the planning agent can not control. 
These actions can not be planned for as it is the case for \i{endogenous} actions, \ie actions which can be executed by the controller.
In our framework, we restrict exogenous actions in that they must have disjoint effect literals.
This is necessary to avoid unwanted side-effects on knowledge which occur due to postdiction. 
Apart from that, we generate the ASP formalisation of an EA $a$ as usual with \hpx\ translation rules but flag it as exogenous by generating the fact $\mathit{ea}(a)$.
In the context of Smart Homes, the motivation behind to considering EA in planning is that ``external'' human agents often intuitively know what to do in a certain situation: 
For instance, if an autonomous wheelchair approaches a person and if the person needs this wheelchair, then it will ``automatically'' sit down on it. 
If human reaction is less automatic, then exogenous actions can often still be triggered by sending appropriate messages to human agents. 
For instance, one can model an exogenous action to fix an abnormality: The controller will notify external maintenance personnel about an abnormality and this should trigger fixing. 
Note that sensing is also allowed as exogenous action.

Though exogenous actions may lead to solutions which would not be found otherwise, the planner should first try to find a plan that does not contain exogenous actions. 
Limiting the number of exogenous actions is realized by the following rules:
{\small
\begin{subequations}
\label{eq:minExo}
\begin{equation}\label{eq:minExo1}
maxExo(N,t) \lam N = @mod(t,\co{n}).
\end{equation}
\begin{equation}\label{eq:numExo1}
numExo(N,t) \lam N = \{occ(A,\_,\_) : ea(A)\}.
\end{equation}
\begin{equation}\label{eq:maxExo1}
\lam maxExo(N,t), numExo(M,t), M > N.
\end{equation}
\end{subequations}
}

Instead of defining an absolute limit, we make the number of allowed exogenous actions dependent on the planning horizon: $\mathit{@mod(t,n)}$\footnote{The clingo family of ASP solvers \cite{Gebser2012} support the definition of \i{lua} functions which can be used for simple auxiliary computation tasks.} returns the modulo of $t$ and a constant $n$ (1st rule), and determines the number of EA that may happen in a certain planning horizon.
The second rule counts the number of exogenous actions and the integrity constraint (third rule) disables stable models where the number of exogenous actions is higher than allowed.\footnote{The integrity constraint appears in the $\mathit{\#volatile}$ part of the program, the other two rules in the $\mathit{\#cumulative}$ part.}

\noindent \f{Explanation} \quad
Where a certain world property may change unexpectedly, it is useful to monitor this property continuously to make sure that the correct value of this property is always known.
For instance, we may open a door and then send a robot through the door. However, we never know whether the door was accidentally closed by another (human) agent in the meanwhile. 

In our framework, unexpected change of world properties is modeled by explanation.
We apply the usual inertia laws and consider unexpected change with abductive explanatory reasoning: 
If a world property changes unexpectedly, then our framework adds the updated knowledge to the domain model indirectly, by considering candidates for exogenous actions that may have caused this change.
We implement this explanation mechanism as follows:
{\small
\begin{equation}
\label{eq:explanation}
\begin{aligned}
&\mathit{0\{exoHappened(A,t,BR)} \\
&~~~~\mathit{	: hasEP(A,EP) : hasEff(EP,\ol{L}) : ea(A)\}  \lam}\\
&~~~~~~~~~~~~~~~\mathit{		knows(L,t,t,B), sensed(\ol{L},t+1)}\\
&~~~~~~~~~~~~~~~\mathit{occ(A,t,BR) \lam exoHappened(A,t,BR).}
\end{aligned}
\end{equation}
}
If it is known that at step $t$ a literal \co{L} holds, but it is sensed that at $t+1$ the complement,  \co{$\ol{L}$} holds, then an exogenous action can have happened that has set \co{L}. 
Note that exogenous actions are only used for explanation if there occurred no endogenous action which may also have set the value of concern: The h-approximation has the restriction that no two actions with the same effect literal may happen concurrently. Therefore, if an endogenous action with the respective effect literal has been executed, an exogenous action with the same effect literal will not be considered for explanation.
Further, explanation relies on the closed world assumption that all actions which can possibly occur are modeled in the domain, and that all exogenous actions have disjoint effect literals. 
Without these requirements it may happen that wrong beliefs are produced:
If there are multiple actions which could explain an unexpected sensing result, then not all explanations will be true. 
If the explanation is wrong, and if the action which is used in the explanation has a condition, then false knowledge about these conditions could be postdicted.
An alternative to the closed-world assumption is to restrict exogenous actions to have only one effect literal and no conditions. 
In that case, even though explanations about the occurrence of actions may be wrong they do not have side-effects on knowledge.
\newpage
\noindent \f{Monitoring} \quad
By monitoring we refer to continuous observation of a world property. As a methodological solution to represent monitoring in the domain specification we suggest to model pseudo-physical effects: If a sensor can monitor the value of a fluent $f$, then we add the physical effect \co{(mon\_$f$)} to the action specification.
For example, the following represents monitoring the open-state of a door:

\begin{lstlisting}
(:action monOpen
	:parameters (?d - Door)
	:effect (mon_open ?d)
	:observe (open ?d))
\end{lstlisting}
Now, to model a ``safe'' drive-action where a door's open state is always known before passing it we add the precondition that the open-state of a door is monitored:


\begin{lstlisting}
(:action drive
		:parameters (?robo - Robot ?door - Door ?from ?to - Room)
		:precondition (and (mon_open ?d)
			$\vdots$
\end{lstlisting}

\subsection{II. Assessing Plan Quality}
So far we have considered ``raw'' weak plans. These plans may still not be very appropriate in practice, \eg because they contain cycles, are unlikely to lead to the goal or involve many exogenous events. 

We use several optimization criteria to asses the quality of plans. An optimal plan should i) contain few exogenous actions ii) achieve the goal for as many contingencies as possible and iii) it should be possible to define soft-constraints or \i{maintenance goals} which must hold whenever possible. Finally, iv) the number of actions should be minimal. 
(With the priority of these criteria in the given order.)

\subsubsection{Plan Strength}
The plan strength reflects for how many contingencies, \ie unknown world properties, the goal is solved. 

\begin{Definition}[{\bf\small Strength of a plan}]
{\small
Given a plan $p$ and a planning problem \mc{P}. Let $n_l(p,\mc{P})$ be the number of leafs of the search tree and $n_g(p,\mc{P})$ the total number of leafs in which the goal is achieved, then the strength $s$ of $p$ \wrt \mc{P} is $s(p,\mc{P}) = \frac{n_g(p,\mc{P})}{n_l(p,\mc{P})}$}
\end{Definition}
In the logic program, the strength for each level $t$ of the conditional plan is determined as follows:
\begin{equation}
\label{eq:planStrength}
\begin{aligned}
\mathit{leafs(L,t) }&\mathit{\lam L=\{uBr(t,B)\}.}\\
\mathit{goals(G,t) }&\mathit{\lam G=\{wGoal(t,B) : uBr(t,B)\}.}\\
\mathit{strength(S,t) }&\lam \\
\mathit{	S=@div}&\mathit{(G*100,L), goals(G,t), leafs(L,t).}
\end{aligned}
\end{equation}
where $L$ is the number of leafs, $G$ is the number of leafs in which the goal is achieved and $S$ is the plan strength. 
\i{oclingo} does only support integer numbers, so $G$ is multiplied by a factor 100 and then divided by $L$.
$\mathit{wGoal}(t,b)$ denotes that all weak goals are achieved in the respective node. 

\subsubsection{Maintenance Goals}
A maintenance goal is a soft-constraint which should hold as often as possible. 
The more nodes a search tree has where a maintenance goal is fulfilled, the higher is its quality. 
We call the corresponding assessment measure the \i{m-value} of a plan $p$ \wrt a planning problem \mc{P}:

\begin{Definition}[{\bf\small m-value of a plan}]{\small
Given a plan $p$ and a planning problem \mc{P}. Let $n_n(p,\mc{P})$ be the number of nodes of the search tree. Let $m_1, \hdots, m_n$ be maintenance goals. Then the m-value $m$ of $p$ \wrt \mc{P} is $m(p,\mc{P}) = \frac{\sum_{m_i}n_{m_i}(p,\mc{P})}{n_n(p,\mc{P})}$, where $n_{m_i}$ is the number of nodes in which a maintenance goal $m_i$ holds.}
\end{Definition}

In terms of Logic Programming, the \emph{m-value} \wrt a planning horizon $t$ is obtained as follows:
{\small
\begin{equation}
\label{eq:countMaintenance}
\begin{aligned}
\mathit{nodes(N,t)} \lam& \mathit{N = {uBr(T,B) : s(T): br(B)}.}\\
\mathit{mSum(M,t) }\lam&  \mathit{M = \{knows(L,t,t,B) :}\\
								&~~~~~~~~~~~~\mathit{uBr(t,B) :  mGoal(L)\}.}\\
\mathit{mVal(V,t) }\lam& \mathit{V=@div(M*100,N),}\\
								&~~~~~~~~~~~~\mathit{mSum(M,t), nodes(N,t).}
\end{aligned}
\end{equation}
}
where  $\mathit{mGoal}(l)$ atoms denote maintenance goals for a literal $l$.

\subsubsection{Applying plan quality measures}
State-of-the-art ASP solvers like \i{clingo} \cite{Gebser2011} offer optimization statements to select an optimal stable model among the entire answer set.
For example, the following statements cause an ASP solver to select one stable model with the minimal number of exogenous actions, maximal strength, maximal m-value and minimal number of actions (with descending priority):
{\small
\begin{equation}
\label{eq:coundMaintenance}
\begin{aligned}
\mathit{\#minimize}& \mathit{[numExo(A,t) = A @ 4]}\\
\mathit{\#maximize}& \mathit{[strength(S,t) = S @ 3]}\\
\mathit{\#maximize}& \mathit{[mVal(M,t) = M @ 2]}\\
\mathit{\#minimize}& \mathit{[act(A,t) = A @ 1]}
\end{aligned}
\end{equation}
}
Unfortunately there is currently no ASP solver available which supports both incremental online problem solving and optimization statements.\footnote{However, this is currently being worked on and a version of \i{oclingo} which supports optimization will be released in near future (\emph{Source}: personal conversation with Torsten Schaub, University of Potsdam).} 
As long as this is not implemented, selecting the plan with the highest quality has to be done by the controller, while the ASP solver can only perform the assessment of the plans. 
However, this is currently not implemented in our prototype.

\section{Case Study: Abnormality Aware\\Wheelchair Robot}
Our framework is integrated into a larger assistance system in a Smart Home environment, namely the Bremen Ambient Assisted Living Lab (BAALL) \cite{Krieg-Bruckner2010}. 
The environment has at its disposal many actuators and sensors such as automatic doors, a smart TV, and also an autonomous robotic wheelchair. 
A typical (simplified) use case in this environment is illustrated in Fig. \ref{fig:onlineCommunication}:
Paul is in the bedroom and wants to get to the bathroom. 
This goal is sent from the controller to the ASP solver, and the planning starts. 
The solver finds the first plan with a horizon of 7 steps. 
The plan is sent to he controller and execution starts.
Door d1 is opened and its open-state is verified by monitoring its open-state. 
It turns out that the door actually is not open and an abnormality is postdicted: $\mathit{knows(ab\_doOpen(d1),0,1,0)}$. 
While these first actions were executed, the solver already incremented the planning horizon up to 10. 
However, to find an alternative this is not sufficient: the route via the office requires 12 steps in total, and accordingly the planning horizon must be at least 12 as well. 
This causes plan execution to be interrupted until the horizon is expanded. 
When the new plan is found, execution continues: Doors d2 and d3 are opened, and their open-states are monitored.
However, while the wheelchair is driving through d2, Mary accidentally closes d3. 
This is immediately reported to the solver: $\mathit{sensed(\neg open(d3),6)}$. 
It interrupts horizon extension to 14, and instead finds an explanation for the closed door -- $\mathit{occ(exoClosed(d3),5,1)}$ -- with a horizon of 13. 
It adopts the plan to the new situation, which considers that d3 has to be opened again. 
Thereafter the rest of the plan can be executed.

\section{Related Work}\label{sec:relWork}
The present work copes with planning, dynamic plan repair, abductive explanation and postdiction under incomplete knowledge and sensing actions. 
Therefore we are interested in other frameworks which have similar features.  
There are many action-theoretic frameworks and implementations such as the Event Calculus Planner by \citeaj{Shanahan2000}, but these often assume complete knowledge about the world and have no semantics that cover sensing actions.

We are interested in formalizations and implementations that cover incomplete knowledge, as found in the literature from the \i{contingent planning} community (\eg CFF \cite{Hoffmann2005} or MBP \cite{Bertoli2001}). 
However these PDDL-based approaches are usually designed for offline-usage and hence not suitable for control tasks as illustrated in our case study. 
In addition, these approaches are usually based on some form of a \pws\ formalization and hence have a higher complexity than \hpx\ (\eg the plan-existence problem for \Ak\ with \pws-semantics is $\Sigma_2^P$ complete). 

PROSOCS \cite{Bracciali2006} is a rich multi-agent framework which supports online reasoning. 
The agents are built according to the KGP model of agency \cite{Kakas2004}. 
The authors use an specialized form of the Event Calculus \cite{Kowalski1986} as reasoning formalism.
PROSOCS supports planning, reactive behavior, goal revision, plan revision and many more features.
Active sensing actions can be specified, but the framework does not support postdiction as part of a contingent planning process: It is not possible to plan for the observation of the effect of an action and then to reason (within the planning) about the condition under which the effect holds. 
Instead, the framework focuses more on multi-agent aspects. 


MAPSIM \cite{Brenner2009} is a continual planning framework based on the planning language MAPL. MAPL is similar to PDDL, but relies on a multi-valued logic. 
In MAPL, the not-knowing of the value of a certain fluent is modeled with a special \co{unknown} value. It is not possible to model conditional effects in MAPL, and hence postdiction is not possible. 
IndiGolog is a high-level programming language by \citeaj{DeGiacomo1998a} which has a search-operator that can also be used to perform planning. 
IndiGolog is capable of planning with incomplete knowledge via a generalized search operator \cite{Sardina2004}. 
However, postdiction and other inference mechanisms have to be implemented by hand, and are thus not elaboration tolerant \cite{McCarthy1998}.

\begin{figure}
\begin{center}
\colorbox{gray!18}{
\begin{minipage}{0.96\columnwidth}
{
\scriptsize
\hfill \textbf{\small ExpCog -- An Experimental Cognitive Robotics Framework}

\smallskip

ExpCog is aimed at integrating logic-based and cognitively-driven agent-control approaches, qualitative models of space and the ability to apply these in the form of planning, explanation and simulation in a wide-range of robotic-control platforms and simulation environments. In addition to its primary experimental function, ExpCog is also geared toward educational purposes. ExpCog provides an easy to use toolkit to integrate qualitative spatial knowledge with formalisms to reason about actions, events, and their effects in order to perform planning and explanation tasks with arbitrary robot platforms and simulators. As demonstrators, support has been included for systems including \emph{ROS}, \emph{Gazebo}, \emph{iCub}. The core integrated agent-control approaches include logic-based approaches like \emph{Situation Calculus}, \emph{Fluent Calculus}, or \emph{STRIPS}, as well as cognitively-driven approaches like \emph{Belief-Desire-Intention}. Furthermore, additional robot platforms and control approaches may be seamlessly integrated.\hspace{0.34in}\hfill \textbf{ExpCog}. \cite{Suchan-Bhatt-expcog-2012}

\textbf{Listing 1}$~$\hfill{\scriptsize\texttt{http://tinyurl.com/expcog}}

}
\end{minipage}}
\end{center}
\end{figure}

\section{Conclusion and Outlook}
We have formalized and implemented an online-planning framework and demonstrated its application in a Smart Home environment. Error-tolerance is achieved by postdicting abnormalities. This requires a formalism like \hpx, which supports sensing along with postdiction. 

On the application side, work is presently in progress to integrate the online h-approximation of this paper within the general experimental cognitive robotics framework ExpCog (Listing 1; \cite{Suchan-Bhatt-expcog-2012}). 
This integration will make is possible for us to release the online planner in a manner such that it may be seamlessly applied for a wide-range for robot control tasks and existing platforms such as ROS ({\small \texttt{http://www.ros.org}}). On the theoretical side we are currently investigating domain-independent heuristics and their formalization in terms of ASP to improve the overall performance of the planner: A huge body of research about heuristics in planning can be found in PDDL-planning related literature, but these heuristics are usually formalized and implemented in procedural formalisms. Transferring these ideas to declarative formalisms such as ASP presents many challenges.

\newpage

\bibliography{PhD,narrative-commonsense-bib}
\bibliographystyle{aaai}

\end{document}